%% file: main.tex
\documentclass[runningheads]{llncs}
\usepackage[T1]{fontenc}
\usepackage{graphicx}
\usepackage{booktabs}
\usepackage[misc]{ifsym}

\usepackage{multirow}
\usepackage{amsfonts}
\usepackage{xspace}
\usepackage{subfigure}
\usepackage{enumitem}
\usepackage{xcolor}
\usepackage{cite}

\usepackage{amsmath}
\usepackage{booktabs}
\usepackage{algorithm}
\usepackage{algorithmic}
\usepackage{threeparttable}
\usepackage{array}

\usepackage{mwe}

\begin{document}

\include{defs}
\title{GLADformer: A Mixed Perspective for Graph-level Anomaly Detection}






\author{Fan Xu \inst{2} \and Nan Wang(\Letter) \inst{1} \and Hao Wu \inst{2} \and Xuezhi Wen \inst{1} \and Dalin Zhang \inst{1} \and Siyang Lu \inst{1} \and Binyong Li \inst{3} \and Wei Gong \inst{2} \and Hai Wan \inst{4}
\and Xibin Zhao(\Letter) \inst{4}}

\authorrunning{F. Xu et al.}

\institute{
School of Software Engineering, Beijing Jiaotong University, Beijing, China \\
\email{\{wangnanbjtu,22126399,dalin,9450\}@bjtu.edu.cn} \\
\and University of Science and Technology of China, Hefei, Chia \\
\email{\{markxu,wuhao2022\}@mail.ustc.edu.cn, weigong@ustc.edu.cn} \\
\and Advanced Cryptography and System Security Key Laboratory of Sichuan Province, Chengdu, China \\
\email{lby@cuit.edu.cn} \\
\and BNRist, KLISS, School of Software, Tsinghua University, Beijing, China \\
\email{\{wanhai,zxb\}@tsinghua.edu.cn}
}

\maketitle              

\begin{abstract}

Graph-Level Anomaly Detection (GLAD) aims to distinguish anomalous graphs within a graph data set. However, current methods are constrained by their receptive fields, struggling to learn global features within the graphs. Moreover, most these methods are based on spatial domain and lack exploration of spectral characteristics. In this paper, we propose a multi-perspective hybrid graph-level anomaly detector named GLADformer, consisting of two key modules. Specifically, we first design a Graph Transformer module with global spectrum enhancement, which ensures balanced and resilient parameter distributions by fusing global features and spectral distribution characteristics. Furthermore, to explore local anomalous attributes, we customize a band-pass spectral GNN message passing module that enhances the model's generalization capability. Through comprehensive experiments on ten real-world datasets from multiple domains, we validate the effectiveness and robustness of GLADformer. This demonstrates that GLADformer outperforms current state-of-the-art graph-level anomaly detection methods, particularly in effectively capturing global anomaly representations and spectral characteristics.

\keywords{Graph-level Anomaly Detection \and Graph Transformer \and Spectral Graph Neural Network.}
\end{abstract}

\input{components/introduction}

\input{components/related_work}

\input{components/methods}

\input{components/experiments}

\section{Conclusion}


In this paper, we rethink the task of graph-level anomaly detection from a multi-perspective view and propose a meticulously designed GLAD model, namely GLADformer. Firstly, we introduce Graph Transformer into the GLAD task to incorporate the inductive bias of graph structures and leverage the differences in spectral energy distribution across graphs. This helps capture global implicit attributes and structural features. Subsequently, departing from the conventional spatial GNNs, we design a novel wavelet spectral GNN for local feature extraction. Further, we propose an improved optimization for the cross-entropy loss function, alleviating the issue of overfitting during training.

\bigskip

\noindent \textbf{Acknowledgement.} This work was supported in part by the Supported by the Fundamental Research Funds for the Central Universities under Grant 2024JBM- C031; in part by the CCF-NSFOCUS Open Fund; in part by the NSFC Program under Grant 62202042, Grant U20A6003, Grant 62076146, Grant 62021002, Grant U19A2062, Grant 62127803, Grant U1911401 and Grant 6212780016; in part by Industrial Technology Infrastructure Public Service Platform Project "Public Service Platform for Urban Rail Transit Equipment Signal System Testing and Safety Evaluation" (No. 2022-233-225), Ministry of Industry and Information Technology of China.

\bibliographystyle{splncs04}
\bibliography{reference}

\end{document}

%% file: defs.tex
\def\A{{\bf A}}
\def\B{{\bf B}}
\def\C{{\bf C}}
\def\D{{\bf D}}
\def\E{{\bf E}}
\def\F{{\bf F}}
\def\G{{\bf G}}
\def\H{{\bf H}}
\def\I{{\bf I}}
\def\J{{\bf J}}
\def\K{{\bf K}}
\def\L{{\bf L}}
\def\M{{\bf M}}
\def\N{{\bf N}}
\def\O{{\bf O}}
\def\P{{\bf P}}
\def\Q{{\bf Q}}
\def\R{{\bf R}}
\def\S{{\bf S}}
\def\T{{\bf T}}
\def\U{{\bf U}}
\def\V{{\bf V}}
\def\W{{\bf W}}
\def\X{{\bf X}}
\def\Y{{\bf Y}}
\def\Z{{\bf Z}}

\def\a{{\bf a}}
\def\b{{\bf b}}
\def\c{{\bf c}}
\def\d{{\bf d}}
\def\e{{\bf e}}
\def\f{{\bf f}}
\def\g{{\bf g}}
\def\h{{\bf h}}
\def\i{{\bf i}}
\def\j{{\bf j}}
\def\k{{\bf k}}
\def\l{{\bf l}}
\def\m{{\bf m}}
\def\n{{\bf n}}
\def\o{{\bf o}}
\def\p{{\bf p}}
\def\q{{\bf q}}
\def\r{{\bf r}}
\def\s{{\bf s}}
\def\t{{\bf t}}
\def\u{{\bf u}}
\def\v{{\bf v}}
\def\w{{\bf w}}
\def\x{{\bf x}}
\def\y{{\bf y}}
\def\z{{\bf z}}
\def\z{{\bf z}}

\def\0{{\bf 0}}
\def\1{{\bf 1}}

\def\AM{{\mathcal A}}
\def\BM{{\mathcal B}}
\def\CM{{\mathcal C}}
\def\HM{{\mathcal H}}
\def\JM{{\mathcal J}}
\def\KM{{\mathcal K}}
\def\PM{{\mathcal P}}
\def\QM{{\mathcal Q}}
\def\WM{{\mathcal W}}
\def\ZM{{\mathcal Z}}
\def\FM{{\mathcal F}}
\def\TM{{\mathcal T}}
\def\UM{{\mathcal U}}
\def\XM{{\mathcal X}}
\def\YM{{\mathcal Y}}
\def\NM{{\mathcal N}}
\def\OM{{\mathcal O}}
\def\IM{{\mathcal I}}
\def\GM{{\mathcal G}}
\def\PM{{\mathcal P}}
\def\LM{{\mathcal L}}
\def\MM{{\mathcal M}}
\def\DM{{\mathcal D}}
\def\SM{{\mathcal S}}
\def\RM{{\mathcal R}}
\def\VM{{\mathcal V}}
\def\EM{{\mathcal E}}

\def\RB{{\mathbb R}}

\def\etal{{\em et al.}}
\def\eg{{\em e.g.}}
\def\ie{{\em i.e.}}
\def\etc{{\em etc}}

\def\Ell{\mbox{\boldmath$\ell$\unboldmath}}
\def\ph{\mbox{\boldmath$\phi$\unboldmath}}
\def\Pii{\mbox{\boldmath$\Pi$\unboldmath}}
\def\pii{\mbox{\boldmath$\pi$\unboldmath}}
\def\Ph{\mbox{\boldmath$\Phi$\unboldmath}}
\def\Ps{\mbox{\boldmath$\Psi$\unboldmath}}
\def\tha{\mbox{\boldmath$\theta$\unboldmath}}
\def\muu{\mbox{\boldmath$\mu$\unboldmath}}
\def\bett{\mbox{\boldmath$\beta$\unboldmath}}
\def\alpp{\mbox{\boldmath$\alpha$\unboldmath}}
\def\Si{\mbox{\boldmath$\Sigma$\unboldmath}}
\def\Gam{\mbox{\boldmath$\Gamma$\unboldmath}}
\def\Lam{\mbox{\boldmath$\Lambda$\unboldmath}}
\def\De{\mbox{\boldmath$\Delta$\unboldmath}}
\def\vps{\mbox{\boldmath$\varepsilon$\unboldmath}}
\def\Lap{\mbox{\boldmath$\LM$\unboldmath}}
\newcommand{\ti}[1]{\tilde{#1}}

\def\tr{\mathrm{tr}}
\def\etal{{\em et al.\/}\,}
\newcommand{\indep}{{\;\bot\!\!\!\!\!\!\bot\;}}
\def\argmax{\mathop{\rm argmax}}
\def\argmin{\mathop{\rm argmin}}
\def\subto{\mathop{\rm s.t.}}
\def\aka{{\em a.k.a.\/}\,}
\def\wrt{{\em w.r.t.\/}\,}
\def\sgn{\mathop{\rm sgn}}

\newtheorem{mydef}{Definition}[section]
\newtheorem{mylem}{Lemma}[section]
\newtheorem{myproof}{Proof}[section]

%% file: components/introduction.tex
\section{Introduction}

Graphs are profoundly employed to model the intricate relationships between data instances across various domains, spanning bioinformatics~\cite{ref1}, chemistry~\cite{ref2}, transportation~\cite{ref3}, and social networks~\cite{ref3}, \etc. Among the downstream tasks for graph data, graph-level anomaly detection~\cite{iGAD} emerges as a significant challenge and encompasses a wide array of application scenarios, such as cancer drug discovery~\cite{ref5} and the identification of toxic molecules~\cite{ref6}.

In recent years, Graph Neural Networks (GNNs) have achieved significant advancements in graph representation learning, like node classification~\cite{GCN}, link prediction~\cite{ref7}, and graph classification~\cite{ref8}, \etc. Specifically, GNNs encode intricate structure and attribute information of graphs into vectors for learning within the latent representation space. To date, a multitude of GNN-based models have been proposed for anomaly detection in graph-structured data~\cite{ref9}, while they predominantly focus on detecting anomalous nodes or edges~\cite{ref10} within a single large graph. In contrast, the domain of graph-level anomaly detection is yet to be extensively explored.

\begin{figure}[!t]
\centering
\includegraphics[width=2.8in]{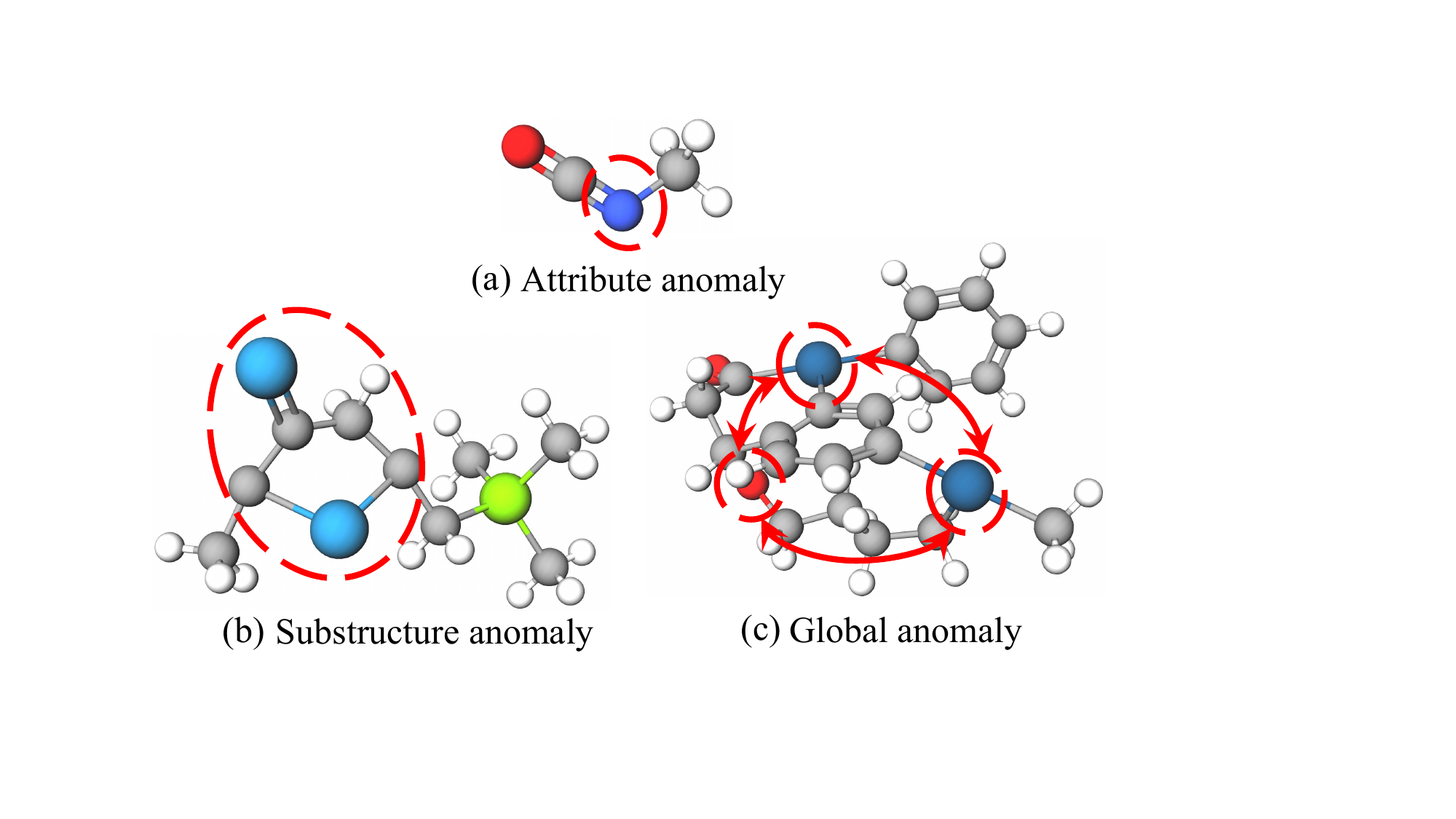}
\caption{Examples for different categories of graph-level anomalies.}
\label{intro}
\end{figure}

Anomalous graphs manifest as outliers and may arise in various scenarios, including local attribute anomalies, substructure anomalies, global interaction anomalies~\cite{GmapAD,HimNet}, \etc. Taking biochemistry molecular data~\cite{Tudataset} as an example, Fig.~\ref{intro} (a) illustrates the toxic molecule of Methyl Isocyanate ($CH_3NCO$), where the nitrogen atom serves as a critical toxic factor. Fig.~\ref{intro} (b) depicts a type of muscarinic molecule, featuring a cyclic peptide structure that distinguishes it from other compounds. Fig.~\ref{intro} (c) displays the main active ingredient in Nux vomica, which is Strychnine ($C_{21}H_{22}N_2O_2$), a toxic ketone alkaloid. However, as Strychnine lacks distinctive elements or specific substructures, researchers need to discern it from the complete molecular structure. Therefore, for a more comprehensive identification of various anomalous graphs, it is imperative to design models that consider both local attributes and structural information while incorporating global interactions within the graph.

The mainstream GLAD models primarily employs GNNs to jointly encode node attributes and topological structural features for acquiring node representations~\cite{iGAD}. Additionally, they design various graph pooling functions~\cite{CVTGAD} to generate graph-level representations for identification of anomalous graphs. However, these methods still face several challenges. On the one hand, traditional GNNs exhibit restricted receptive fields~\cite{ref16}, focusing solely on local neighbors or subgraph information of the current node, thereby lacking the ability to capture long-distance information interactions and global features. Although some methods attempt to learn intra-graph and inter-graph knowledge by maintaining a repository of anomalous node or graph candidates~\cite{GmapAD}, they still fall short in capturing the global information within the graphs. On the other hand, when learning local features, spatial-domain GNNs tend to overlook essential high-frequency information and underlying semantic details due to their low-pass filtering characteristics~\cite{ref15}. Moreover, the current graph-level pooling mechanisms exhibit constrained generalization capabilities, leading to erratic performances across disparate datasets.

To address the aforementioned challenges, we propose a novel graph-level anomaly detection model, which combines spectral-enhanced global perception and local multi-frequency information guidance. Our method, GLADformer, primarily consists of two key modules: the Spectrum-Enhanced Graph Transformer module and the Local Spectral Message Passing module. Specifically, we first introduce a Graph Transformer operator in the spatial domain, where effective graph-induced biases such as node degrees and structural information are jointly inputted, incorporating explicit spectral distribution deviations to capture anomaly information from a global perspective. Subsequently, to better explore local features and mitigate the limitations of spatial GNNs, we design a novel wavelet spectral GNN to learn discriminative local attributes and structural features in a complex spectral domain. Finally, to overcome class imbalance issues and excessive confidence of traditional cross-entropy loss, we propose an improved variation-optimized cross-entropy loss function. Our main contributions can be summarized as follows:


\begin{itemize}
    \item[$\bullet$] Our approach not only incorporates a Graph Transformer module designed in the spatial domain but also integrates spectral energy distribution deviations to enhance global perception. Additionally, we design a spectral GNN with multi-frequency message passing characteristics to guide the extraction of local anomaly features.
    \item[$\bullet$] To better alleviate the issue of class imbalance and overcome the limitations of using cross-entropy as a measure for anomaly detdection, we propose a weighted variation-optimized cross-entropy loss function.
    \item[$\bullet$] Comprehensive experiments on a variety of datasets across ten baselines demonstrate that GALDformer exhibits competitive performance in terms of both effectiveness and robustness.
\end{itemize}


%% file: components/related_work.tex
\section{Related Work}

\subsection{Graph-Level Anomaly Detection}

In recent years, graph anomaly detection has garnered extensive attention across various domains. However, most existing approaches focus on detecting anomalous nodes or edges within an individual graph~\cite{xu2024revisiting,10476155}, while graph-level anomaly detection remains largely unexplored. 

GLAD aims to differentiate deviant structures or abnormal properties within a single graph to identify anomalous graphs that exhibit substantial differences compared to the majority in a collection. State-of-the-art end-to-end methods leverage powerful GNN backbones and incorporate advanced strategies to learn graph representations suitable for anomaly detection. For instance, GLocalKD~\cite{GLocalKD} and OCGTL~\cite{OCGTL} respectively combine GNN with knowledge distillation and one-class classification to detect anomalous graphs. iGAD~\cite{iGAD} introduces an abnormal substructure-aware deep random walk kernel and a node-aware kernel to capture both topology information and node features. To better explore inter-graph information, GmapAD~\cite{GmapAD} maps individual graphs to a representation space by computing similarity between graphs and inter-graph candidate nodes, achieving high discriminability between abnormal and normal graphs. Some recent works focus on interpretable analysis of graph-level anomaly detection and have achieved promising results. For instance, SIGNET~\cite{SIGNET} measures the anomaly degree of each graph based on cross-view mutual information~\cite{CMSCGC} and extracts bottleneck subgraphs in a self-supervised manner to provide explanations for anomaly discrimination. However, existing methods mostly originate from a spatial perspective, and there is still a lack of exploration regarding the influence of graph-level spectrum energy information.

\subsection{Graph Transformer}

Transformer~\cite{Transformer,PastNet} has achieved overwhelming advantages in the NLP~\cite{Bert, wu2024slfnet} and CV~\cite{ViT, liu2021swin} domains, and recently many researchers have been devoted to extending Transformer to the study of graph-structured data~\cite{ref14}. One of the strengths of Transformer is its ability to capture global receptive fields, but it lacks the capability to capture positional information, which poses significant limitations in graph data~\cite{2024AMGC}. Recently, researchers investigate the use of Position Encoding (PE) and Structure Encoding (SE)~\cite{ref19,ref20} within the graph domain to capture various types of graph structure features, leveraging techniques such as shortest path proximity~\cite{ref23} or spectral information to enhance inductive bias. For example, Graphformer~\cite{ref14} designs novel structural position encoding that outperforms popular GNN models in a wide range of graph prediction tasks. Further, SAN~\cite{ref21} employs both sparse and global attention mechanisms at each layer and introduces learnable Position Encoding (LapPE) to replace static Laplacian eigenvectors. Exphormer~\cite{ref22} explores sparse attention mechanisms with virtual global nodes and extended graphs, showcasing linear complexity and desirable theoretical properties.

%% file: components/methods.tex
\section{Preliminary}

\subsection{Problem Definition}

Given a graph set $\hat{\mathcal{G}}=\{G_1,G_2,...,G_N\}$, each graph is denoted as $G_i=(\mathcal{V}_i,\mathcal{E}_i)$, where $\mathcal{V}_i$ is the set of nodes and $\mathcal{E}_i$ is the set of edges. The node features can be represented as $X_i \in R^{N_i \times d}$, and the edge information can be denoted as an adjacency matrix $A_i \in [0,1]^{N_i\times N_i}$. In this paper, we concentrate on supervised graph-level anomaly detection, thus we aim to learn an anomaly labeling function based on the given training graphs and their graph labels $y_{i}=\{0,1\}$. Then we hope to assign a high anomaly score to a graph $G$ in the testing set if it significantly deviates from the majority.

\subsection{Graph Spectrum and Rayleigh Quotient}

For each graph $G$ in the graph set, the adjacency matrix is denoted by $A$. The diagonal degree matrix $D$ is defined as $(D)_{ii}=\sum_j(A)_{ij}$, and the normalized laplacian matrix $L$ is defined as $I-D^{-\frac{1}{2}}AD^{-\frac{1}{2}}$. The laplacian $L$ can be decomposed into its eigenvectors and eigenvalues as $U\Lambda U^T$. The diagonal elements of $\Lambda$ are composed of its eigenvalues: $\Lambda=\operatorname{diag}([\lambda_1,\lambda_2,\ldots,\lambda_N])$, and the eigenvalues satisfy $0 \le \lambda_{1}\le\ \cdot\cdot\cdot \le \lambda_{N} \le 2$. Let $X=\left (x_{1},x_{2},...,x_{N} \right )$ be a signal on graph $G$, then $U^T X$ is the graph Fourier transformation of $X$. Then we introduce Rayleigh quotient to demonstrate the standardized variance fraction of signal $X$. 

\begin{equation}
    R(L,X) 
    = \frac{X^{T}LX}{X^{T}X} 
    = \frac{\sum_{i=1}^{N} \lambda_{i} \tilde{x}_{i}^2}{\sum_{i=1}^{N} \tilde x_{i}^2}
    = \lambda_{N} * f(t)_{max} - S_{spec},
\end{equation}

\noindent where $S_{spec}$ denotes the the integration of signal energy in the frequency domain (with respect to eigenvalues)~\cite{BWGNN}. Thus Rayleigh quotient can also denote the contribution of high-frequency region.

\section{Method}

\begin{figure}[!t]
\centering
\includegraphics[width=4.8in]{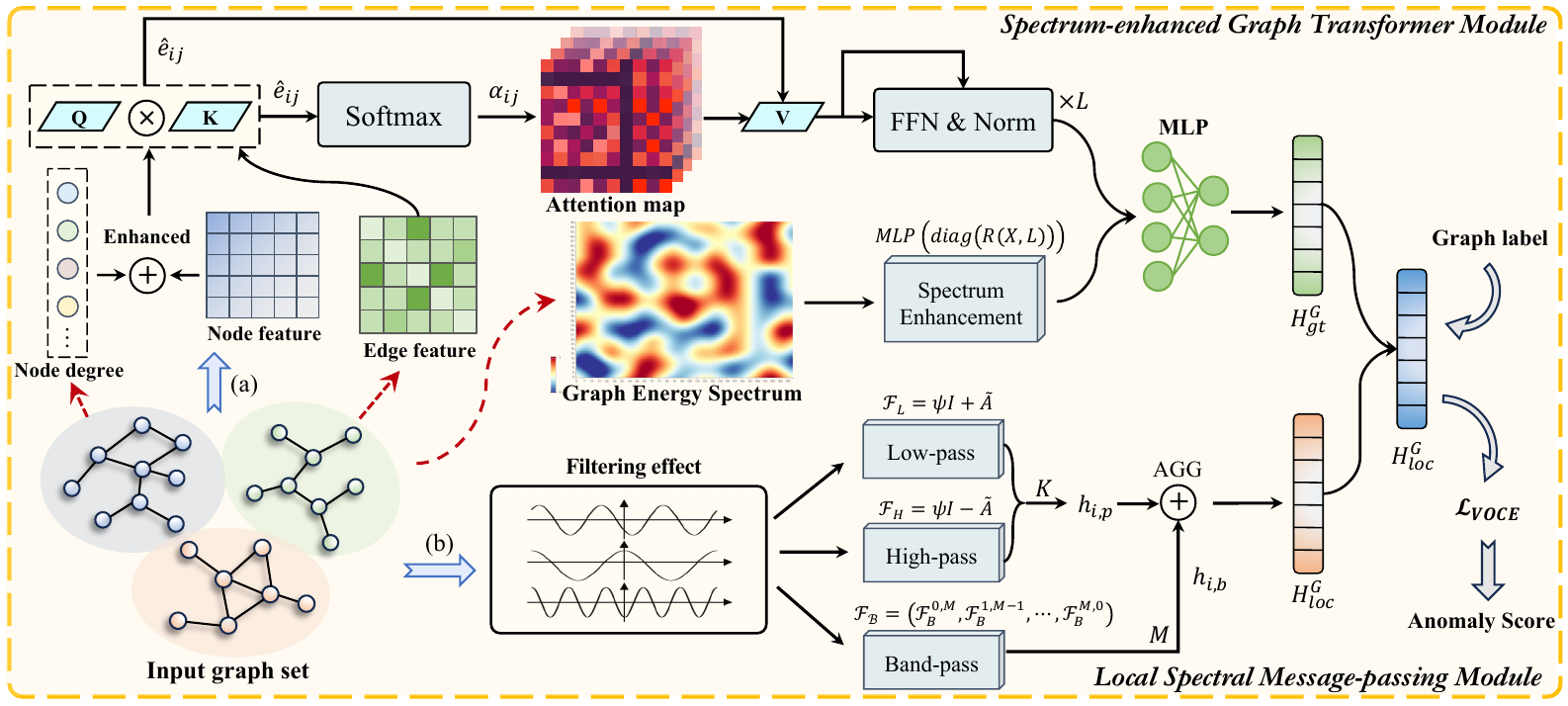}
\caption{The above image presents the overview of our model GLADformer, where (a) and (b) demonstrate the the spectrum-enhanced graph-transformer module and the local spectral message passing module respectively.}
\label{framework}
\end{figure}



In this section, we will systematically describe the technical details of the GLADformer framework (view Fig.~\ref{framework} for demonstration), which consists of three core components. Specifically, we first detail the design of our spectrum-enhanced of Graph Transformer from a global perspective (Section \ref{sec4.1}). Subsequently, we explore the discriminative local attribute and structure features through a multi-frequency spectral GNN (Section \ref{sec4.2}). Finally, we present the meticulously designed variation-optimize cross-entropy loss function (Section \ref{sec4.3}).

\subsection{Spectrum-enhanced Graph Transformer Module} \label{sec4.1}



Initially, we establish a super-node in each graph, which forms connections with all remaining nodes within that graph. Further, to address the issue of the Transformer architecture's lack of strong inductive biases and to leverage the advantages of message passing strategy in GNNs, we incorporate appropriate graph structures and relative position encoding. We first employ a single-layer MLP to obtain the initial node representations.

\begin{equation}
    h_i = \phi(W x_i + b).
\end{equation}

Subsequently, we enhance the node features by introducing an adaptive degree-scaler to maintain the degree information.

\begin{equation}
    \mathsf{g}_i=h_i\cdot\theta_1+\left(\log(1+deg_{i})\cdot h_i\cdot\theta_2\right),
\end{equation}

\noindent where ${\theta}_1, {\theta}_2\in R^d$ are learnable weights, and $deg_i$ is the degree of node $i$. Then we utilize Relative Random Walk Probabilities (RRWP)~\cite{ref25} to initialize the relative position encoding, thus capturing intriguing graph structure information. According to~\cite{ref23}, RRWP has been proved to be more expressive than shortest path distances (SPD)~\cite{ref26} through recently proposed Weisfeiler-Leman-like graph isomorphism tests~\cite{ref27}. Let $A$, $D$ be the adjacency matrix and degree matrix of a graph, we can define $R := AD^{-1}$, and $R_{ij}$ denotes the possibility of node $i$ transitioning to node $j$ in one step.


Then the RRWP initial positional encoding for each pair of nodes is illustrated as $R_{i,j}=[I,R,R^2,\ldots,R^{T-1}]_{i,j} \in R^T$. To make full use of the structural information, we utilize the position encoding $R_{ij}$ as the edge feature $e_{ij}$. Further, we cleverly combine Transformer's self-attention mechanism and customized edge features to design a novel Graph Transformer architecture. Specifically, in the stage of calculating attention score, we replace the scalar product of Query and Key with the vector product~\cite{yan2024dual}, and inject the initialized edge features to obtain learnable edge features. This learnable edge feature is subsequently utilized in the calculation of attention score and output encoding. The specific formula is as follows:

\begin{equation}
    \begin{aligned}
    \hat{e}_{ij}&=\sigma((\mathsf{g}_{i}W_{Q}\otimes \mathsf{g}_{j}W_{K})+e_{ij}W_{E}),\\
    \alpha_{ij}&=Softmax(\hat{e}_{ij}\cdot W_{\alpha}),\\
    \hat{\mathsf{g}}_{i}&=\sum_{j\in V}\alpha_{ij}\left(\mathsf{g}_{j} W_{V} + \hat{e}_{ij} \right),
    \end{aligned}
\end{equation}

\noindent where $W_Q, W_K, W_V \in R^{d \times d^{\prime}}$, $W_{E} \in R^{K \times d^{\prime}}$ and $W_{\alpha}\in R^{d^{\prime} \times 1}$ are learnable weight matrices; and $\otimes$ denotes element-wise multiplication of vectors. 

Similar to other Transformer architectures, we can easily convert the aforementioned attention mechanism into multi-head attention. Then we follow~\cite{ref23} to obtain the output of the $l$-th layer, denoted as $\mathcal{G}^{(l)}$:

\begin{equation}
    \begin{aligned}
    \mathcal{G}^{^{\prime}(l)}&=\text{Norm}(\text{RRWP-MHA}(\mathcal{G}^{(l-1)})+\mathcal{G}^{(l-1)}),\\
    \mathcal{G}^{(l)}&=\text{Norm}(\text{FFN}(\mathcal{G}^{^{\prime}(l)}))+\mathcal{G}^{^{\prime}(l)},
    \end{aligned}
\end{equation}

\noindent where $\text{Norm}(\cdot)$ indicates layer-norm function, $\text{RRWP-MHA}$ is the multi-head attention mechanism designed before, and $\text{FFN}$~\cite{Transformer} is a feed-forward network. Afterward, we merge the encoding representations from each layer of the super-node to obtain the graph-level representations.

\begin{equation}
    H^{G}_{sup}=f_{\mathrm{COMBINE}}\left(\mathcal{G}_{sup}^{(l)}\mid l=1,...,L\right).
\end{equation}



\subsubsection{Spectrum Enhancement.}

Subsequently, as discussed in section 3.2, we have already known the positive correlation between Rayleigh quotient and high-frequency components. According to~\cite{RQ}, the spectral energy distributions of normal and anomalous graphs are distinct in the graph spectral domain. Therefore, it is crucial to incorporate spectral energy distributions into the GLAD task. However, accurately calculating the ratios of spectral energies requires the eigenvalue-eigenvector decomposition of the graph Laplacian matrix, which has a complexity of $O(n^3)$ and is computationally expensive for large-scale graphs. To avoid high time complexity, we employ the Rayleigh quotient $R(L,X)=\frac{X^T L X }{X^TX}$ as a substitute. For signal $X \in R^{N \times d}$, it can be proven that the diagonal elements of the Rayleigh quotient $R(L,X)$ correspond to the spectral characteristics of the signal $X$ in each feature dimension on the graph. 

Given the potential sparsity of graph adjacency matrices and graph signals, as well as the possibility of high discretization levels, we utilize a single-layer MLP to derive the Rayleigh quotient vector for each graph. This vector denotes the explicit Rayleigh quotient feature and it primarily focuses on the global spectral distribution characteristics of the graph. Then, we utilize this to enhance the previously obtained graph-level representation.




\begin{equation}
    \begin{aligned}
    H_{rq}^G &= \text{MLP} \left(diag \left (R(X,L) \right ) \right ),    \\
    H_{gt}^G =& \text{MLP}  \left( \text{CONCAT} \left( H^{G}_{sup},H_{rq}^G \right) \right ).
    \end{aligned}
\end{equation}

\subsection{Local Spectral Message-passing Module} \label{sec4.2}

To devise an adaptive local message passing filter, we introduce the Beta-distributed wavelet basis~\cite{BWGNN}, which conforms to the Hammond's graph wavelet theory~\cite{ref28} and is band-pass in nature. The underlying probability distribution function is $f_{Beta}\left ( w \right ) = w^\alpha(1-w)^\beta / B(\alpha+1,\beta+1)$, where $w\in \left [ 0, 1 \right ]$, and $B(\alpha+1,\beta+1)=\alpha!\beta!/(\alpha+\beta+1)!$ is a constant. According to~\cite{BWGNN}, since the eigenvalues of the normalized Laplacian matrix satisfy $\lambda \in \left [ 0,2 \right ] $, we utilize a variation like $f_{Beta}^{\ast}(\lambda)=\frac{1}{2}f_{Beta}(\frac{\lambda}{2})$. The obtained band-pass filter is expressed below:

\begin{equation}
    \mathcal{F}_{B}^{\alpha,\beta}=U f_{Beta}^{\ast }(\Lambda)U^T =\frac{(\frac L2)^{\alpha}(I-\frac L2)^{\beta}}{2B(\alpha+1,\beta+1)}.
\end{equation}

\noindent Mapping this band-pass wavelet kernel into the graph spectral domain, $M = \alpha + \beta$ denotes the neighboring receptive field order of this filter. In turn, we obtain the spectral domain Beta wavelet transform group $\mathcal{F_B}$:

\begin{equation}
    \mathcal{F_B}=(\mathcal{F}_B^{0,M},\mathcal{F}_B^{1,M-1},\cdotp\cdotp\cdotp,\mathcal{F}_B^{M,0}).
\end{equation}

Contrasting with conventional graph neural network message passing mechanisms, this work employs parallel wavelet kernels for message propagation. Subsequently, it combines the corresponding filtered information.

\begin{equation}
    h_{i,b}=f_{\mathrm{COMBINE}}\left(\mathcal{F}_B^{m,M-m}\cdotp h_i^{(0)}\mid m=1,...,M\right),
\end{equation}

\noindent where $h_{i,b}$ denotes the node representation of $v_i$ after band-pass filtering. 

Upon meticulous analysis, it is ascertained that each constituent of the Beta band-pass wavelet kernel manifests as an amalgam of elevated powers of the adjacency matrix $A$ and the Laplacian matrix $L$. This architecture conspicuously lacks specialized low-pass and high-pass filtering kernels imperative for mitigating lower and higher frequencies, potentially precipitating the omission of pivotal information. To rectify this shortfall, we propose the adoption of the generalized Laplacian low-pass and high-pass filters: 

\begin{equation}
    \begin{aligned}
    \mathcal{F}_L&=(\psi+1)I-L=U[(\psi+1)I-\Lambda]U^\top,\\
    \mathcal{F}_H&=(\psi-1)I+L=U[(\psi-1)I+\Lambda]U^\top,
    \end{aligned}
\end{equation}

\noindent where $\psi \in \left [0,1 \right ]$ plays a pivotal role in modulating the characteristics of both low-pass and high-pass filter kernels. Subsequently, the dual filter kernels are concurrently utilized to discern and assimilate the pure low-frequency and high-frequency signals aggregating from neighbors, thereby enhancing the model's fidelity to local structures.

\begin{equation}
    \begin{aligned}
    &h_{i,p}^{(l)}=\text{MLP} \left ( \text{AGG} \left ( \mathcal{F}_L \cdotp h_i^{(l-1)}, \mathcal{F}_H \cdotp h_i^{(l-1)} \right ) \right ) ,\\
    &h_{i,p}=f_{\mathrm{COMBINE}}\left(h^{(l)}_{i,p}\mid l=1,...,K\right).
    \end{aligned}
\end{equation}

\noindent The features from each layer are sequentially concatenated and subsequently amalgamated to obtain the corresponding aggregated low-order neighbor attributes. Ultimately, the feature is amalgamated with the band-pass filtering outcomes, procuring the definitive local graph-level representation.

\begin{equation}
    H^{G}_{loc}= \text{READOUT}\left ( \text{AGG} \left (h_{i,b}, h_{i,p} \right ) \mid v_i\in G \right ),
\end{equation}

\noindent where $\text{READOUT}$ function can be achieved by permutation invariant graph pooling functions like summation or mean~\cite{GmapAD}. Finally, we fuse the features obtained from two modules to obtain the final graph-level representation:

\begin{equation}
H^G=\text{MLP}\left(\text{CONCAT}\left(H_{gt}^G,H_{loc}^G\right)\right).
\end{equation}

\subsection{Variation-optimize Cross-entropy Loss Function} \label{sec4.3}


The cross-entropy loss is widely used in various classification tasks and is often employed as the loss function for designing anomaly detection. However, cross-entropy suffers from phenomena such as overconfidence and struggles to adapt to scenarios like data imbalance. It is well known that the conventional evaluation metric for classification problems is accuracy, but cross-entropy is not a smooth approximation of accuracy, which significantly impacts the precision of model predictions. For instance, when the predicted probability of training samples is very low, cross-entropy tends to yield a tremendously large loss, even though these data points are likely to be noises. This phenomenon is particularly pronounced in anomaly detection data, leading to overfitting of the model to noise data. To address these issues and cater to the graph anomaly detection scenarios, we propose a novel cross-entropy loss function.


Approaching from a gradient-based perspective, if accuracy is employed as the evaluation metric, the gradient with respect to $p_\theta(y|x)$ is $-\nabla_\theta p_\theta(y|x)$. Conversely, for the cross-entropy $-\log p_\theta(y|x)$, the gradient is $-\frac{1}{p_\theta(y|x)}\nabla_\theta p_\theta(y|x)$ ($y_{true}$ is not involved in the gradient computation and has been omitted). Now, we construct a new gradient based on these two equations:

\begin{equation}
    -{\frac{1}{\kappa+(1-\kappa)p_\theta(y|x)}} \hspace{0.5cm} \kappa \in \left [ 0, 1 \right ] .
\end{equation}

This new gradient retains the advantages of cross-entropy while synchronizing better with changes in accuracy, thus overcoming the problem of overfitting. Subsequently, we seek the original function based on this differential gradient.

\begin{equation}
    -\frac{\nabla_\theta p_\theta(y|x)}{\kappa+(1-\kappa)p_\theta(y|x)}=\nabla_\theta\left(-\frac{\log\left[\kappa+(1-\kappa)p_\theta(y|x)\right]}{1-\kappa}\right).
\end{equation}

By incorporating the original function into the GLAD task, we can obtain the variation-optimize cross-entropy loss function $\mathcal{L}_{VOCE}$.

\begin{equation}
\mathcal{L}_{VOCE} = -\partial y\frac{\log\left[\kappa+(1-\kappa)p\right]}{1-\kappa}-(1-y)\frac{\log\left[\kappa+(1-\kappa)(1-p)\right]}{1-\kappa},
\end{equation}

\noindent where the hyperparameter $\kappa$ is utilized to modulate the balance between accuracy and cross-entropy within the loss function ($\kappa$ is set to 0.2 in experiments), and $\partial$ is logarithm of the ratio between normal and anomalous samples.

%% file: components/experiments.tex
\section{Experiment}

In this section, we perform thorough experiments to validate the effectiveness of our proposed GLADformer method against nine baselines on ten real-world datasets. Furthermore, we conduct comprehensive ablation tests on GLADformer and perform visual analysis of key components. 


\subsection{Experiment Settings}

\noindent \textbf{Datasets.}
We conduct experiments on 10 real-world graph datasets from two popular application domains: (\romannumeral1) BZR, AIDS, COX2, NCI1, ENZYMES, and PROTEINS are collected from biochemistry~\cite{Tudataset}, samples from minority or truly abnormal classes are considered as anomalies. Following~\cite{GmapAD}, the selected anomaly samples are downsampled, retaining only 10\% of the samples. (\romannumeral2) MCF-7, MOLT-4, SW-620, and PC-3 are collected from PubChem~\cite{PubChem}, and they reflect the anti-cancer activity test results of a large number of compounds on cancer cell lines. Chemical compounds that exhibit antibody activity against cancer are labeled as abnormal graphs, while others are labeled as normal graphs. The statistics of these graph datasets are summarized in Table~\ref{tab:dataset}.


\begin{table}[h]\scriptsize
  \centering
   \tabcolsep=1.70mm
  \caption{The statistics of the 10 datasets.}
  \label{tab:dataset}
    \begin{tabular}{cccccccc}
    \toprule
    \textbf{Dataset} & \textbf{$N.G$} & \textbf{$N.A$} & \textbf{$Ratio\%$} & \textbf{$AVG.n$} & \textbf{$AVG.e$} & \textbf{$Attr$}  \\
    \hline
    AIDS   & 2000 & 400 & 20.00  &15.69  &16.20  &4   \\
    BZR   & 405 & 86 & 21.23  &35.75  &38.36  &3   \\
    COX2   & 467 & 102 & 21.84  &41.22  &43.45  &3   \\
    NCI1  & 4110 & 2053  & 49.95   &29.87  &32.30  & 37   \\
    ENZYMES & 600 & 100  & 16.67   &32.63  &62.14  &18   \\
    PROTEINS & 1113 & 450   &40.43  &39.06  &72.82  &29  \\
    MCF-7 & 25476 & 2294  & 8.26  &26.39  &28.52  &46   \\
    MOLT-4 & 36625 & 3140  & 7.90   &26.07  &28.13  &64   \\
    SW-620 & 38122 & 2410  & 5.95   &26.05  &28.08  &65   \\
    PC-3  & 25941 & 1568  & 5.70   &26.35  &28.49  &45   \\
    \bottomrule
    \end{tabular}%
\end{table}

\begin{table*}[!t]
\centering
\scriptsize
\renewcommand{\arraystretch}{1.2} 
\setlength\tabcolsep{4.5pt} 
\caption{The performance comparison in terms of AUC value (in percent, mean value). Best result in bold, second best underlined.}
\label{tab:auc}
\begin{threeparttable}
\begin{tabular}{l|p{0.6cm}<{\centering} p{0.6cm}<{\centering} p{0.6cm}<{\centering}|p{0.8cm}<{\centering} p{0.6cm}<{\centering}|p{0.9cm}<{\centering} p{0.6cm}<{\centering} p{0.7cm}<{\centering} p{0.9cm}<{\centering}|p{0.7cm}<{\centering}}
\toprule
Datasets & \tiny GCN & \tiny GAT & \tiny GIN & \tiny SAGPool & \tiny GMT & \tiny GLocalKD & \tiny iGAD & \tiny HimNet & \tiny GmapAD & \tiny Ours \\
\hline
BZR & 55.87 & 57.37 & 58.92 & 61.78 & 65.91 & 67.95 & 69.37 & 70.38 & \underline{72.58} & \textbf{77.25}  \\ 
AIDS & 86.37 & 88.29 & 87.62 & 90.48 & 92.16  & 93.24 & 97.62 & 98.71 & \underline{99.06} & \textbf{99.62}  \\ 
COX2 & 50.21 & 52.77 & 52.08 & 54.38 & 53.58 & 58.93 & 61.48 & \underline{63.76} & 62.73 & \textbf{68.47}  \\ 
ENZYMES & 53.94 & 55.61 & 54.75 & 56.90 & 58.75  & \underline{63.27} & 60.92 & 58.94 & 57.62 & \textbf{63.86}  \\ 
PROTEINS & 68.70 & 69.53 & 70.05 & 72.58 & 74.94  & 76.41 & 75.93 & 77.28 & \textbf{78.35} & \underline{77.68}  \\ 
NCI1 & 62.34 & 64.28 & 63.98 & 66.32 & \underline{71.98} & 68.38 & 70.45 & 68.63 & 71.52 & \textbf{76.83}  \\ 
MCF-7 & 64.48 & 65.26 & 65.62 & 71.64 & 77.06  & 63.63 & \underline{81.46} & 63.69 & 71.28 & \textbf{83.58}  \\ 
PC-3 & 66.97 & 67.36 & 67.95 & 69.37 & 78.96  & 67.27 & \textbf{85.63} & 67.03 & 74.26 & \underline{84.19}  \\
MOLT-4 & 63.74 & 65.21 & 64.82 & 65.11 & 76.06  & 66.31 & \underline{82.79} & 66.33 & 69.82 & \textbf{83.24}  \\  
SW-620 & 59.87 & 62.31 & 61.30 & 72.51 & 74.67  & 65.42 & \underline{84.86} & 65.44 & 72.97 & \textbf{85.73}  \\ 
\bottomrule
\end{tabular}
\end{threeparttable}
\end{table*}

\begin{table*}[!t]
\centering
\scriptsize
\renewcommand{\arraystretch}{1.2} 
\setlength\tabcolsep{4.5pt} 
\caption{The performance comparison in terms of F1 score (in percent, mean value). Best result in bold, second best underlined.}
\label{tab:f1}
\begin{threeparttable}
\begin{tabular}{l|p{0.6cm}<{\centering} p{0.6cm}<{\centering} p{0.6cm}<{\centering}|p{0.8cm}<{\centering} p{0.6cm}<{\centering}|p{0.9cm}<{\centering} p{0.7cm}<{\centering} p{0.7cm}<{\centering} p{0.9cm}<{\centering}|p{0.7cm}<{\centering}}
\toprule
Datasets & \tiny GCN & \tiny GAT & \tiny GIN & \tiny SAGPool & \tiny GMT & \tiny GLocalKD & \tiny iGAD & \tiny HimNet & \tiny GmapAD & \tiny Ours \\
\hline
BZR &     51.27 & 52.31  & 52.97  & 55.93  & 54.35 & 56.12 & 56.74 &  58.14 &  \underline{59.82} & \textbf{61.87} \\ 
AIDS &    68.10 & 70.83  & 71.64  & 74.28  & 74.32 & 76.63 & 77.93 &  78.92 &  \underline{80.17} & \textbf{82.47} \\ 
COX2 &    43.28 & 43.75  & 45.15  & 47.33  & 48.85 & 50.82 & 51.35 &  \underline{52.18} &  51.38 & \textbf{57.05} \\ 
ENZYMES & 44.73 & 43.95  & 44.16  & 47.18  & 46.94 & \underline{50.23} & 48.02 &  48.85 &  48.53 & \textbf{53.28} \\ 
PROTEINS& 49.82 & 49.26  & 50.08  & 55.74  & 58.75 & 61.03 & 60.61 &  52.47 &  \textbf{62.84} & \underline{61.74} \\ 
NCI1 &    51.17 & 50.84  & 51.20  & 59.03  & \underline{61.27} & 57.92 & 58.09 &  56.93 &  58.29 & \textbf{64.29} \\ 
MCF-7 &   48.83 & 47.97  & 48.05  & 58.83  & 62.12 & 60.31 & \underline{64.68} &  57.58 &  60.47 & \textbf{65.84} \\ 
PC-3 &    48.79 & 47.98  & 48.56  & 59.65  & \underline{63.87} & 59.93 & 67.10 &  61.25 &  63.31 & \textbf{67.37} \\ 
MOLT-4 &  49.87 & 50.32  & 50.02  & 59.37  & 62.07 & 61.74 & \underline{66.81} &  60.53 &  62.84 & \textbf{68.06} \\ 
SW-620 &  48.65 & 48.84  & 48.74  & 58.60  & 61.28 & 60.86 & \underline{66.23} &  59.46 &  61.73 & \textbf{68.58} \\ 
\bottomrule
\end{tabular}
\end{threeparttable}
\end{table*}

\noindent \textbf{Comparison Method.} To illustrate the effectiveness of our proposed model, we conduct extensive experiments between GLADformer and 9 competitive methods, which can be classified into three groups. (\romannumeral1) Spatial GNNs with average pooling function: GCN~\cite{GCN}, GAT~\cite{GAT} and GIN~\cite{GIN}. (\romannumeral2) Graph classification methods: SAGPool~\cite{SAGPool} and GMT~\cite{GMT}. (\romannumeral3) State-of-the-art deep GLAD methods: GLocalKD~\cite{GLocalKD}, iGAD~\cite{iGAD}, HimNet~\cite{HimNet}, and GmapAD~\cite{GmapAD}.

\noindent \textbf{Evaluation Metrics.} We evaluate methods using popular anomaly detection metrics, AUC values and Macro-F1 scores~\cite{AUC}. The results are reported by performing 5-fold cross-validation for all datasets.

\noindent \textbf{Experimental Settings.} For our model, the dimensions of the hidden layers and output features for all three modules are set to 128 and 32 respectively. During training, the hyperparameters $T, L, M, K$ are set to 4, 6, 3 and 4 separately, and the batch size is 128 for all datasets. We use Adam as the optimizer with a learning rate of 0.001 and utilize cross-entropy loss as the basic loss function. Each dataset is randomly shuffled and split for training, validation, and testing with ratios of 70\%, 15\%, and 15\%. For GCN, GAT and GIN, we utilize 2 layers and apply average pooling function to obtain graph-level representations. For other baselines, we use their published settings unless the parameters are specially identified in the original paper. All experiments in this work are conducted on an NVIDIA A100-PCIE-40GB.


\subsection{Main Results}

We first compare GLADformer with the aforementioned baselines of different categories. The overall performances of all methods with respect to AUC value and F1 score against ten datasets are illustrated in Table~\ref{tab:auc} and Table~\ref{tab:f1}. Among all the ten datasets, GLADformer achieves the highest AUC values in eight and the highest F1 scores in nine. Specifically, we have the following observations:



\begin{itemize}
    \item When compared with the spatial GNNs combined with average graph pooling function, GLADformer demonstrates significant advantages on ten real world datasets in terms of AUC value and F1 score. The performance of conventional GCN, GAT, and GIN models falls short of expectations, this may be because they can only capture local low-frequency features, and the average pooling function further exacerbates the over-smoothing issue. 

    \item Subsequently, when compared with the two graph classification methods, SGAPool and GMT obtain some performance improvement compared to traditional methods, which can be attributed to their specially designed pooling strategies. However, they are not specifically tailored for detecting anomalous graphs, and GLADformer explores more comprehensive global and local information and fully considers spectral characteristics, thus there still exists a considerable gap between them and GLADformer.

    \item Finally, in comparison with the state-of-the-art GLAD methods, our approach achieves superior performance on nearly all datasets in terms of AUC value and F1 score. Specifically, the performance improvement over OCGTL and GLocalKD demonstrates the effectiveness of considering global and local features in intra-graph collaboration and exploring spectral characteristics within and between graphs for graph-level anomaly detection. Moreover, the HimNet and GmapAD methods achieve the modest performance on the first six biochemistry datasets, indicating the effectiveness of maintaining and updating a pool of anomalous nodes or subgraphs on certain datasets with attribute anomalies. It is noteworthy that iGAD achieves commendable performance on the four Pubchem datasets. We hypothesize that this may be because the iGAD method encodes rich local substructure information into the graph-level embeddings, and these four datasets contain a multitude of substructure anomalies. The aforementioned results validate the superiority and effectiveness of GLADformer in graph-level anomaly detection tasks. 
\end{itemize}

\subsection{Ablation Study}


To evaluate the performance impact of different components on our proposed model GLADformer, we conduct comprehensive ablation study experiments. Specifically, for the two key modules: spectrum-enhance graph-transformer module (\textbf{GT}) and local spectral message-passing module (\textbf{LS}), we construct a total of five variant models, and the overall results are demonstrated in Table~\ref{tab:ablation}. From top to bottom, the five variant models represent the following: (1) Elimination of the GT module, (2) Elimination of the spectral enhancement component within the GT module, (3) Elimination of the LS module, (4) Replacement of the LS module with GIN, and (5) Replacement of the LS module with BernNet~\cite{BernNet}.

\begin{table}[!h]
\centering
\scriptsize
\caption{Experimental results of ablation with different model variants on four datasets with respect to AUC value. Best result in bold, second best underlined.}
\label{tab:ablation}
\begin{threeparttable}
\begin{tabular}{l p{1.2cm}<{\centering} p{1.2cm}<{\centering} p{1.2cm}<{\centering} p{1.5cm}<{\centering}}
\toprule
\textbf{Model} & \textbf{BZR} & \textbf{NCI1} & \textbf{MCF-7} & \textbf{MOLT-4} \\
\midrule
GLADformer w/o GT & 72.56 & 71.30 & 77.64 & 78.47 \\
GLADformer (GT w/o SEC) & \underline{76.40} & \underline{76.18} & \underline{82.37} & 81.92 \\
GLADformer w/o LS  & 73.68 & 72.20 & 78.19 & 78.70 \\
GLADformer w/o LS + GIN & 73.73 & 72.56 & 78.23 & 79.18 \\
GLADformer w/o LS + BernNet & 75.68 & 74.79 & 80.85 & \underline{82.03} \\
\midrule
GLADformer & \textbf{77.25} & \textbf{76.83} & \textbf{83.58} & \textbf{83.24} \\
\bottomrule
\end{tabular}
\end{threeparttable}
\end{table}




As illustrated in Tables~\ref{tab:ablation}, we observe that removing either the GT module or the LS module resulted in a significant decline in model performance, with the performance decrease being more pronounced when the GT module was eliminated. Furthermore, when the spectrum enhancement component (SEC) within the GT module is removed, we notice a slight performance degradation. This indicates that the distribution of spectral energy indeed plays a guiding role in determining the anomalous properties of the graph. 
Additionally, when replacing the LS module with different local GNNs like GIN and BernNet, we find the model performs better with the use of BernNet, which possesses spectral filtering characteristics. This suggests that the low-pass filtering effect smooths out the local feature, resulting in the losses of local anomaly information.

\subsection{Visualization Analysis}

\begin{figure}[!t]
\centering
\includegraphics[width=4.8in]{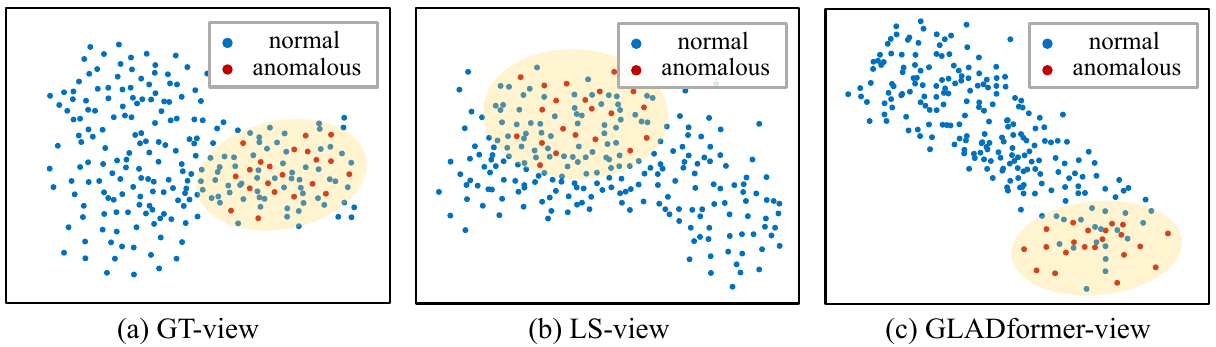}
\caption{Visualization analysis on NCI1 dataset.}
\label{sandian}
\end{figure}

To further demonstrate the performance of our model and validate the effectiveness of each module, we conduct extensive visual analysis in this subsection. We firstly load the parameters of the last two models that only retain a single module in the ablation experiment, as well as the complete GLADformer model. Specifically, we employ t-SNE to visualize the embeddings generated by these three models. Due to space constraints, we present visualization results for the NCI1 dataset only in this section, the visual results are presented in Fig.~\ref{sandian}. 

As observed in Fig.~\ref{sandian}, the three figures respectively represent the embeddings output of individual GT module, LS module, as well as the entire GLADformer model, visualized in a two-dimensional space. We observe that each of the first two figures demonstrates a certain level of discriminability between anomalous and normal graphs, verifying the meaningfulness of the two core modules. Then in the third figure, the anomalous graphs are well separated from the normal graphs, and they exhibit distinct distributions. This observation highlights the successful integration of the two modules we designed, further confirming their synergistic effectiveness.







